\documentclass[10pt]{article}

\usepackage{fullpage}
\usepackage{setspace}
\usepackage{parskip}
\usepackage{titlesec}
\usepackage[section]{placeins}
\usepackage{xcolor}
\usepackage{breakcites}
\usepackage{lineno}
\usepackage{hyphenat}
\usepackage{microtype}
\usepackage[edges]{forest}
\usepackage[T1]{fontenc}
\usepackage{wrapfig}
\usetikzlibrary{fit}
\usepackage{caption}

\PassOptionsToPackage{hyphens}{url}
\usepackage[colorlinks = true,
            linkcolor = blue,
            urlcolor  = blue,
            citecolor = blue,
            anchorcolor = blue]{hyperref}
\usepackage{etoolbox}
\makeatother

\usepackage{natbib}

\renewenvironment{abstract}
  {{\bfseries\noindent{\abstractname}\par\nobreak}\footnotesize}
  {\bigskip}

\titlespacing{\section}{0pt}{*3}{*1}
\titlespacing{\subsection}{0pt}{*2}{*0.5}
\titlespacing{\subsubsection}{0pt}{*1.5}{0pt}

\usepackage{authblk}

\usepackage{graphicx}
\usepackage[space]{grffile}
\usepackage{latexsym}
\usepackage{textcomp}
\usepackage{longtable}
\usepackage{tabulary}
\usepackage{booktabs,array,multirow}
\usepackage{amsfonts,amsmath,amssymb}
\usepackage{tikz}
\usetikzlibrary{shapes,positioning}

\providecommand\citet{\cite}
\providecommand\citep{\cite}

\newif\iflatexml\latexmlfalse

\AtBeginDocument{\DeclareGraphicsExtensions{.pdf,.PDF,.eps,.EPS,.png,.PNG,.tif,.TIF,.jpg,.JPG,.jpeg,.JPEG}}

\usepackage[utf8]{inputenc}
\usepackage[english]{babel}

\usepackage{float}

\usepackage[margin=1.5in]{geometry}
\usepackage{tabularx}

\begin{document}

\title{Why are NLP Models Fumbling at Elementary Math? \\ A Survey of Deep Learning based Word Problem Solvers}

\author[1]{Sowmya S Sundaram}%
\author[2]{Sairam Gurajada}%
\author[1]{Marco Fisichella}%
\author[3]{Deepak P}%
\author[4]{Savitha Sam Abraham}%
\affil[1]{L3S Research Center, Hannover, Germany}%
\affil[2]{IBM Research, Almaden, USA\footnote{Work done while author was here}}%
\affil[3]{Queen's University, Belfast, UK}%
\affil[4]{University of Örebro, Sweden}%

\vspace{-1em}

  \date{}

\begingroup
\let\center\flushleft
\let\endcenter\endflushleft
\maketitle
\endgroup

\selectlanguage{english}
\begin{abstract}
From the latter half of the last decade, there has been a growing interest in developing algorithms for automatically solving mathematical word problems (MWP). It is a challenging and unique task that demands blending surface level text pattern recognition with mathematical reasoning. In spite of extensive research, we are still miles away from building robust representations of elementary math word problems and effective solutions for the general task. In this paper, we critically examine the various models that have been developed for solving word problems, their pros and cons and the challenges ahead. In the last two years, a lot of deep learning models have recorded competing results on benchmark datasets, making a critical and conceptual analysis of literature highly useful at this juncture. We take a step back and analyse why, in spite of this abundance in scholarly interest, the predominantly used experiment and dataset designs continue to be a stumbling block. From the vantage point of having analyzed the literature closely, we also endeavour to provide a road-map for future math word problem research. 
\end{abstract}%

\sloppy

\section*{Introduction}

Natural language processing has been one of the most popular and intriguing AI-complete sub-fields of artificial intelligence. One of the earliest systems arguably was the PhD Thesis on automatically solving arithmetic word problems \cite{bobrow1964question}. The challenge lay on two fronts (a) analysing unconstrained natural language, and (b) mapping intricate text patterns onto a small mathematical vocabulary, for usage within its reasoning framework.


Right up until 2010, there has been prolific exploration of MWP solvers, for various domains (such as algebra, percentages, ratio etc). These solvers relied heavily on hand-crafted rules for bridging the gap between language and the corresponding mathematical notation. As can be surmised, these approaches, while being effective within their niches, did not generalise well to address the broader problem of solving MWPs. Moreover, due to the lack of well accepted datasets, it is hard to measure the relative performance across proposed systems \cite{mukherjee2008review}.

\begin{table}[htb!]
    \centering
    \footnotesize
    \def\arraystretch{1.5}
    \begin{tabular}{|l p{5cm}|}
    \hline
        \textbf{Input} & Kevin has 3 books. Kylie has 7 books. How many books do they have together? \\
        \textbf{Answer} & 10 \\
        \hline
    \end{tabular}
    \caption{Typical Example}
    \label{tab:example}
\end{table}

The pioneering work by \cite{kushman2014learning} employed statistical methods to solve word problems, which set the stage for the development of automatic MWP solvers using traditional machine learning methods. The work also introduced the first dataset, popularly referred to as Alg514, that had multiple linear equations associated with a problem. The machine learning task was to map the coefficients in the equation to the numbers in the problem. The dataset comprises data units with a triplet structure: natural language question, equation set, and the final answer.

Mirroring recent trends in NLP, there has been an explosion of deep learning models for MWP. Some of the early ones~\cite{wang2017deep, ling2017program} modeled the task of converting the text to equation as a sequence-to-sequence (seq2seq, for short) problem. In this context, increasingly complex models have been proposed to capture semantics beyond the surface text. Some have captured structural information (pertaining to input text, domain knowledge, output equation structure) in the form of graphs and used advances in graph neural networks (\cite{li-etal-2020-graph-tree}, \cite{zhang-etal-2020-graph-tree}, etc.). Others have utilised the benefits of transformers in their modelling (\cite{mwp-bert}, \cite{piekos-etal-2021-measuring}, etc.). We will explore these models in detail.

Since this is a problem that has consistently attracted steady (arguably, slow and steady) attention, ostensibly right from the birth of the field of NLP, a survey of the problem solving techniques offers a good horizon for researchers. The authors collected 30+ papers on deep learning for word problem solving, published over the last three years across premier NLP avenues. Each paper has its own unique intuitive basis, but most achieve comparable empirical performance. The profusion of methods has made it hard to crisply point out the state-of-the-art, even for fairly general word problem solving settings. Hence, a broad overview of the techniques employed gives a good grounding for further research. Similarly, understanding the source, settings and relevance of datasets is often important. For example, there are many datasets that are often referred to by multiple names at different points in time. Also, the finer aspects of problem scenario varies across systems (whether multiple equations can be solved, whether it is restricted to algebra or more domains etc.). In this survey, we systematically analyse the models, list the benchmark datasets and examine word problem solving literature using a critical analysis perspective.





\subsection*{Related Surveys}

There are two seminal surveys that cover word problem solving research. One,~\cite{mukherjee2008review}, has a detailed overview of the symbolic solvers for this problem. The second, more recent one \cite{survey2}, covers models proposed up until 2020. In the last two years, there has been a sharp spike in algorithms developed, that focus on various aspects of deep learning, to model this problem. Our survey is predominantly based on these deep learning models. The differentiating aspects of our survey from another related one, \cite{other-survey} are: the usage of a critical perspective to analyze deep learning models, which enables us to identify robustness deficiencies in the methods analytically, and also to trace them back to model design and dataset choice issues. We will also include empirical performance values of various methods on popular datasets, and deliberate on future directions.


\section*{Symbolic Solvers}
We begin our discussion with traditional solvers that employ a rule-based method to convert text input to a set of \textit{symbols}. Early solvers within this family such as STUDENT \cite{bobrow1964question} and other subsequent ones (\cite{fletcher1985understanding}, \cite{dellarosa1986computer}), the dominant methodology was to map natural language input to an underlying pre-defined \textit{schema}. This calls for a mechanism to distil common expectations of language, word problems and the corresponding mathematical notation, to form bespoke rulesets that will power the conversion. This may be seen as setting up a slot-filling mechanism that map the main entities of the word problem to a slots within a set of equation templates. 

An example of a schema for algebraic MWP is shown in Table~\ref{tab:symbsolver}.

\begin{table}[h]
    \centering
    \footnotesize
    \def\arraystretch{1.5}
    \begin{tabular}{|p{1.7cm} p{4.3cm}|}
    \hline 
    \textbf{Problem} & John has 5 apples. He gave 2 to Mary. How many does he have now? \\
    \hline 
    \textbf{Template} & [Owner$_1$] has [X] [obj]. \\
    & [Owner$_1$] [transfer] [Y] [obj] to [Owner$_2$]. \\
    & [Owner$_1$] has [Z] [obj]. \\
    & Z = X - Y \\
    \hline 
    \textbf{Slot-Filling} & [John] has [5] [apple]. \\
    & [John] [give] [2] [apple] to [Mary]. \\
    & [Mary] has [Z] [apple]. \\
    & Z = 5 - 2 \\
    \hline 
    \textbf{Answer} & Z = 3 \\
    \hline
    \end{tabular}
    \caption{Workflow of Symbolic Solvers}
    \label{tab:symbsolver}
\end{table}

The advantage is that these systems are robust in handling irrelevant information, with expert-authored rulesets enabling focus towards pertinent parts of the problem. To further enhance the practical effectiveness within applications focusing niche domains, research focused on tailoring these symbolic systems for target domains~\cite{mukherjee2008review}. As one can observe, the rules would need to be exhaustive to capture the myriad nuances of language. Thus, they did not generalise well across varying language styles. Since each system was designed for a particular domain, comparative performance evaluation was hindered by the unavailability of cross-domain datasets. 
 

\section*{Statistical Solvers}

As with many tasks in natural language processing, statistical machine learning techniques to solve word problems started dominating the field from 2014. The central theme of these techniques has been to score a number of potential solutions (may be equations or expression trees as we will see shortly) within an optimization based scoring framework, and subsequently arrive at the correct mathematical model for the given text. This may be thought of as viewing the task as a \textbf{structure prediction challenge} \cite{survey2}.

\begin{equation}
P(y|x; \theta) = \frac{e^{\theta.\phi(x,y)}}{\sum_{y' \in Y}e^{\theta.\phi(x,y')}}
\end{equation}

As with optimization problems, Equation 1 refers to the problem of learning parameters $\theta$, which relate to the feature function $\phi$. Consider labeled dataset $D$ consisting of $n$ pairs $(x, y, a)$ where $x$ is the natural language question, $y$ is the mathematical expression and $a$ is the numerical answer. The task is to score all possible expressions $Y$, and maximise the choice of the labelled $y$ through an optimisation setting. This is done by modifying the parameters $\theta$ of the feature function $\phi(x,y)$. Different models propose different formulations of $\phi$. In practise, beam search is used as a control mechanism. We grouped the prolific algorithms that were developed, based on the type of mathematical structure $y$ - either as \textit{equation templates} or \textit{expression trees}. Equation templates were mined from training data, much like the slot filling idea of symbolic systems. However, they became a bottleneck to generalizability, if the word problem at inference time, was from an unseen equation template. To address this issue, expression trees, with unambiguous post-fix traversals, were used to model equations. Though they restricted the complexity of the systems to single equation models, they offered wider scope for generalizability. 


\subsection*{Equation Templates}
Equation templates extract out the numeric coefficients and maintain the variable and operator structure. This was used as a popular representation of mathematical modelling.
To begin with, \cite{kushman2014learning}, used structure prediction to score both equation templates and alignment of the numerals in the input text to coefficients in the template. Using a state based representation, \cite{hosseini2014} modelled simple elementary level word problems with emphasis on verb categorisation. \cite{zhou2015learn} enhanced the work done by \cite{kushman2014learning} by using quadratic programming to increase efficiency. \cite{upadhyay-chang-2017-annotating} introduced a sophisticated method of representing derivations in this space. 

\subsection*{Expression Trees}
Expression trees are applicable only to single equation systems. The single equation is represented as a tree, with leaves of the tree being numbers and the internal nodes being operators as illustrated in \cite{koncel2015parsing}.

Expression tree based methods converge faster, understandably due to the diminished complexity of the model. Some solvers (such as \cite{roy2016solving}) had a joint optimisation objective to identify relevant numbers and populating the expression tree. On the other hand, \cite{koncel2015parsing, mitra2016learning} used domain knowledge to constrain the search space.

\section*{Neural Solvers}

Among the major challenges for the solvers we have seen so far was that of converting the input text into a meaningful feature space to enable downstream solving; the main divergences across papers seen across the previous sections has been based on the technological flavour and methodology employed for such text-to-representation conversion. 

\begin{figure}[!ht]
\centering
\scriptsize
\begin{forest}
  for tree={
    draw,
    text width=1.4cm,
    align=center
  },
  forked edges,
  [Automatic \\ Word \\ Problem \\ Solvers
    [Symbolic \\ Solvers]
    [Statistical \\ Solvers
        [Expression \\ Trees]
        [Equation \\ Templates]
    ]
    [Neural \\ Solvers
        [Seq2Seq]
        [Graph-Based]
        [Transformers]
        [Contrastive]
        [Knowledge \\ Distillation]
    ]
  ]
  \node [draw, fit=(current bounding box.south east) (current bounding box.north west)] {};
\end{forest}

  \captionof{figure}{Types of Word Problem Solvers}
    \label{fig:my_label}
\end{figure}
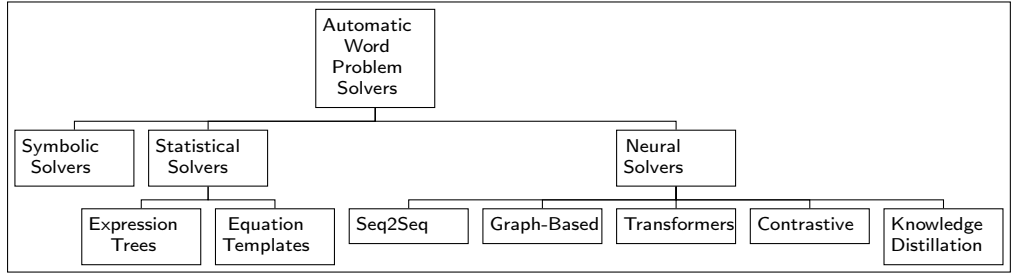

The advent of distributed representations for text~\cite{le2014distributed, elmo, glove, DBLP:journals/corr/abs-1810-04805}, marked a sharp departure in the line of inquiry towards solving math word problems, focusing on the details of the learning architecture rather than feature-space modelling. There have even been domain specific distributed representation learners for word problems~\cite{sundaram2020distributed}. As an example of solvers, \cite{ling2017program} designed a seq2seq model that incorporated learning a program as an intermediate step. This and other early works made it fashionable to treat the word problem solving task as a language translation task, i.e., translating from the input natural language text to a sequence of characters representing either the equation or a sequence of predicates. This design choice, however, has its limitations, which are sometimes severe in terms of the restrictions they place on math problems that can be admitted within such architectures \cite{patel-etal-2021-nlp}. A few of these linguistic vs. math structure understanding challenges, especially for neural solvers, are illustrated in Table ~\ref{tab:challenges}. As an important example, equation systems that involve solving multiple equations are not straightforward to address within such a framework. A notable exception to this is the popular baseline MathDQN \cite{Wang_Zhang_Gao_Song_Guo_Shen_2018}, which employs deep reinforcement learning. We consider different families of deep learning solvers within separate sub-sections herein.

\begin{table}[b]
    \centering
    \footnotesize
    \def\arraystretch{1.5}
    \begin{tabular}{|l p{5cm}|}
    \hline
        \textbf{Input} & Kevin has \colorbox{cyan}{3} books. Kylie has \colorbox{cyan}{7} books and 3 pencils. How many books do they have together? \\
        \textbf{Mathematical Structure} & \colorbox{cyan}{3} + \colorbox{cyan}{7} \\
        \textbf{Linguistic Structure} & (Person1) has (X) (object1). (Person2) has (Y) (object1) and (Z) (object2). \\
        \hline
        \textbf{Challenges} & (1) Order of X and Y does not matter in addition (2) multiple equations do not make a sequence (3) Similar objects need to be grouped together  \\
        \hline
    \end{tabular}
    \caption{Typical Challenges}
    \label{tab:challenges}
\end{table}


\subsection*{Seq2Seq Solvers}
The ubiquitous Seq2Seq \cite{seq2seq} architecture is widely popular for automatic word problem solving. From early direct use of LSTMs \cite{Hochreiter:1997} / GRUs \cite{cho-etal-2014-learning} in Seq2Seq models (\cite{huang2017learning}, \cite{wang2017deep}) to complex models that include domain knowledge \cite{ling2017program, qin-etal-2020-semantically, chiang-chen-2019-semantically, qin-etal-2021-neural}), diverse formulations of this basic architecture have been employed.

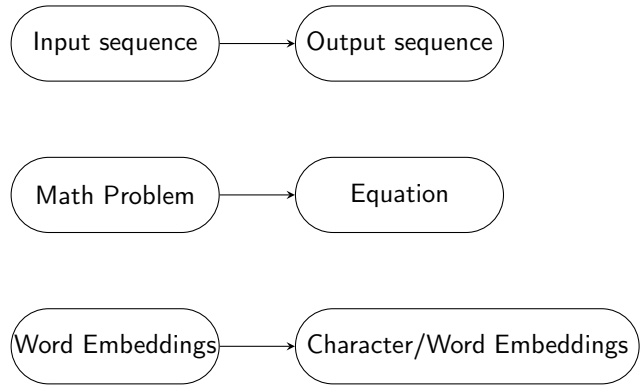
\begin{figure}[ht!]
\centering
\begin{tikzpicture}
    \node[draw,
    rounded rectangle,
    minimum width=3cm,
    minimum height=1cm] (block1) {Input sequence};
    \node[draw,
    right=of block1,
    rounded rectangle,
    minimum width=3cm,
    minimum height=1cm] (block2) {Output sequence};
    \node[draw,
    below=of block1,
    rounded rectangle,
    minimum width=3cm,
    minimum height=1cm] (block3) {Math Problem};
    \node[draw,
    right=of block3,
    rounded rectangle,
    minimum width=3cm,
    minimum height=1cm] (block4) {Equation};
    \node[draw,
    below=of block3,
    rounded rectangle,
    minimum width=3cm,
    minimum height=1cm] (block5) {Word Embeddings};
    \node[draw,
    right=of block5,
    rounded rectangle,
    minimum width=3cm,
    minimum height=1cm] (block6) {Character/Word Embeddings};
    \draw[-stealth] (block1) -- (block2);
    \draw[-stealth] (block3) -- (block4);
    \draw[-stealth] (block5) -- (block6);
\end{tikzpicture}

  \captionof{figure}{General Seq2Seq Formulations}
    \label{fig:seq2seq}
\end{figure}

The initial set of models used Seq2Seq as is, with small variations in the usage of LSTM or GRUs or with simple heuristics (for example, \cite{huang-etal-2016-well} used retrieval to enhance the results). Significant improvements were made by including some mathematical aspects. This, once again, demonstrates that the task is not merely that of language translation. \cite{ling2017program} converted the word problem to a text containing the explanation or rationale. This was done through an intermediate step of generating a step-by-step program on a large dataset. Though the accuracy values reported were low, the domains spanned anywhere between probability to relative velocity, and the unified framework demonstrated performing meaningful analysis through qualitative illustrations. This was improved upon by \cite{amini-etal-2019-mathqa}, which enhanced the dataset and added domain information through a label on the category. The SAU-Solver \cite{qin-etal-2020-semantically} introduced a tree like representation with semantic elements that align to the word problem. As seen in Table ~\ref{tab:performance}, this is a formidable contender. In \cite{chiang-chen-2019-semantically}, a novel way of decomposing the equation construction into a set of stack operations - such that more nuanced mapping between language and operators can be learned - was designed. There is a burgeoning section of the literature that is invested in using neuro-symbolic reasoning to bridge this gap between perception level tasks (language understanding) and cognitive level tasks (mathematical reasoning). An example of this is \cite{qin-etal-2021-neural}. With this discussion, it is clear that adding some form of domain knowledge benefits an automatic solver.          

\subsection*{Graph-based Solvers}

With the advent of graph modeling \cite{xia2019graph} and enhanced interest in multi-modal processing, the graph data structure became a vehicle for adding knowledge to solvers. One way of enabling this has been to simply model the input problem as a graph~\cite{feng-etal-2021-graphmr,li-etal-2020-graph-tree,yu-etal-2021-improving,hong2021learning}. This incorporates domain knowledge of (a) language interactions pertinent to mathematical reasoning, or (b) quantity graphs stating how various numerals in the text are connected. Another way is to model the decoder side to accept graphical input of equations \cite{tree-ijcai2019,hms,expert-systems,Cao_Hong_Li_Luo_2021,liu-etal-2019-tree,wu-etal-2021-math}. Another natural pathway that has been employed towards leveraging graphs is to use graph neural networks for both encoder and decoder \cite{zhang-etal-2020-graph-tree,wu-etal-2020-knowledge,wu-etal-2021-edge-enhanced,shen-jin-2020-solving}.

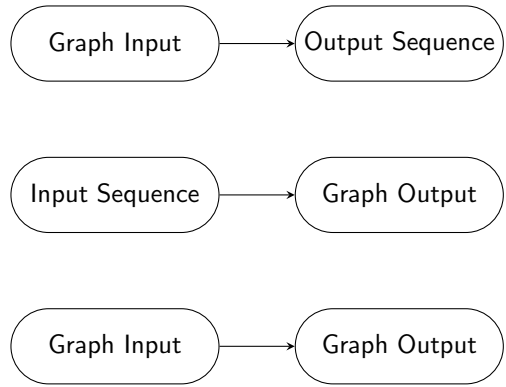
\begin{figure}[ht!]
\centering
\begin{tikzpicture}
    \node[draw,
    rounded rectangle,
    minimum width=3cm,
    minimum height=1cm] (block7) {Graph Input};
    \node[draw,
    right=of block7,
    rounded rectangle,
    minimum width=3cm,
    minimum height=1cm] (block8) {Output Sequence};
    \node[draw,
    below=of block7,
    rounded rectangle,
    minimum width=3cm,
    minimum height=1cm] (block9) {Input Sequence};
    \node[draw,
    right=of block9,
    rounded rectangle,
    minimum width=3cm,
    minimum height=1cm] (block10) {Graph Output};
    \node[draw,
    rounded rectangle,
    below=of block9,
    minimum width=3cm,
    minimum height=1cm] (block11) {Graph Input};
    \node[draw,
    right=of block11,
    rounded rectangle,
    minimum width=3cm,
    minimum height=1cm] (block12) {Graph Output};
    \draw[-stealth] (block7) -- (block8);
    \draw[-stealth] (block9) -- (block10);
    \draw[-stealth] (block11) -- (block12);
    
    \end{tikzpicture}

  \captionof{figure}{General Graph based Formulations}
    \label{fig:graph-based}
\end{figure}

Graphs are capable of representing complex relationships. With the time-tested success of graph neural networks (GNNs) \cite{gnn-review}, they fit easily into the encoder-decoder architecture. Intuitively, when graphs are used on the input side, we can model complex semantic relationships in the linguistic side of the task. When graphs are used on the decoder side, relationships between the numerical entities or an intermediate representation of the problem can be captured. Analogously, graph-to-graph modelling enables matching the semantics of both language and math. This does not necessarily imply graph-to-graph outperforms all the other formulations. There are unique pros and cons of each of the graph-based papers, as both language and mathematical models are hard to (a) model separately and (b) model the interactions. The interesting observation as seen in Table ~\ref{tab:performance}, graph based models are both popular and powerful. Unlike sequences, when the input text is represented as a graph, the focus is more on relevant entities rather than a stream of text. Similarly quantity graphs or semantics informed graphs, eliminate ordering ambiguities in equations. This formulation, however, still does not address the multiple equation problem.

\subsection*{Transformers} 
Transformers \cite{DBLP:journals/corr/VaswaniSPUJGKP17} have lately revolutionised the field of NLP. Word problem solving has been no exception. Through the use of BERT \cite{DBLP:journals/corr/abs-1810-04805} embeddings or through transformer based encoder-decoder models, some recent research has leveraged concepts from transformer models \cite{liu-etal-2019-tree,kim-etal-2020-point}. The translation has been modeled variously, such as from text to explanation \cite{piekos-etal-2021-measuring,griffith-kalita-2020-solving}, or from text to equation \cite{shen-etal-2021-generate-rank,mwp-bert}.

When moving from Word2Vec \cite{word2vec} vectors to BERT embeddings \cite{DBLP:journals/corr/abs-1810-04805}, massive gains were expected due to (a) greater incorporation of context level information and (b) automatic capturing of relevant information as BERT is essentially a Masked Language Model. Interestingly, the gains do not have as large a margin as seen in other language tasks such as question answering or machine translation \cite{DBLP:journals/corr/abs-1810-04805}. BERT is a large model that needs to be fine tuned with domain specific information. The small gains point towards low quality of word problem datasets, which is in line with the fact that the datasets are either quite small by deep learning standards or that they have high lexical overlap, effectively suggesting that the set of characteristic word problems are small.

\subsection*{Contrastive Solvers}
    
With the widespread usage of Siamese networks~\cite{koch2015siamese}, the idea of building representations that \textit{contrast} between vectorial representations across classes in data has seen some interest. In the context of word problem solving, a few bespoke transformer based encoder-decoder models \cite{contrastive,hong2021learning} have been proposed; these seek to effectively leverage contrastive learning~\cite{ContrastiveLearning2020}.

This is a relatively new paradigm and more research needs to emerge to ascertain definite trends. One of the main stumbling blocks of word problem solving is that two highly \textit{linguistically} similar looking word problems may have entirely different \textit{mathematical} structure. Since contrastive learning is built on the principle that similar input examples lead to closer representations, it allows one to use the notion of similarity and dissimilarity to overcome this bottleneck and consciously design semantically informed intermediate representations, such that the similarity is built not only from the language vocabulary, but also from the mathematical concepts. 

\subsection*{Teacher-Student Solvers}

The paradigm of knowledge distillation, in the wake of large, generic end-to-end models, has become popular in NLP \cite{knowledge-distillation}. The underlying idea behind this is to \textit{distill} smaller task-specific models from a generic large pre-trained or generic model. Since word problem datasets are of comparatively smaller size, it is but logical that large generic networks can be fine-tuned for downstream processing of word problem solving, as favourably demonstrated by \cite{ijcai2020-555} and \cite{hong2021learning}.

Once again, this is an emerging paradigm. Similar to the discussion we presented with transformer based models, the fact that the presence of pre-trained language models alone is not sufficient for this task has bolstered initial efforts in this direction. Knowledge distillation enables a model to focus the learnings of one generic model on to a smaller, more focussed one, especially with less datapoints. Hence, the method of adding semantic information through the usage of knowledge distillation algorithms is promising and one to look out for. 

\section*{Domain-Niche Solvers}

Some research, encompassing families of statistical solvers and deep models, focus on the pertinent characteristics of a particular domain in mathematics, such as probability word problems \cite{dries2017solving, suster-etal-2021-mapping, tsai-etal-2021-sequence}, number theory word problems \cite{shi2015automatically}, geometry word problems \cite{seo2015solving,chen-etal-2021-geoqa} and age word problems \cite{sundaram2019semantic}. 

\section*{Datasets}

Datasets used for math word problem solving are listed in Table~\ref{tab:datasets} with their characteristics. The top section of the table describes datasets with relatively fewer data objects ($\leq 1k$, to be specific). The bottom half consists of more recent datasets that are larger and more popularly used within deep learning methodologies. 

\begin{table}[t]
    \centering
    \def\arraystretch{1.1}
    \begin{tabularx}{\textwidth}{l l l l l}
    \hline
    \textbf{Dataset} & \textbf{Type} & \textbf{Domain} & \textbf{Size} & \textbf{Source} \\
    \hline\hline
        Alg514  & Multi-equation & (+,-,*,/) & 514 & \cite{kushman2014learning}\\
        \scriptsize{(SimulEq-S)} &&&& \\
        AddSub & Single-equation & (+,-) & 340 & \cite{hosseini2014} \\
        \scriptsize{(AI2)} &&&& \\
        SingleOp  & Single-equation & (+,-,*,/) & 562 & \cite{roy2015reasoning} \\
        \scriptsize{(Illinois, IL)} &&&& \\
        SingleEq & Single-equation & (+,-,*,/) & 508 & \cite{koncel2015parsing} \\
        MAWPS & Multi-equation & (+,-,*,/) & 3320 & \cite{mawps} \\
        MultiArith  & Single-equation & (+,-,*,/) & 600 & \cite{roy2016solving} \\
        \scriptsize{(Common Core, CC)} &&&& \\
        AllArith & Single-equation & (+,-,*,/) & 831 & \cite{roy_solving} \\
        Perturb & Single-equation & (+,-,*,/) & 661 & \cite{roy_solving} \\
        Aggregate & Single-equation & (+,-,*,/) & 1492 & \cite{roy_solving} \\
        DRAW-1k & Multi-equation & (+,-,*,/) & 1k & \cite{upadhyay-chang-2017-annotating} \\
        AsDIV-A & Single-equation & (+,-,*,/) & 2373 & \cite{asdiva} \\
        SVAMP & Single-equation & (+,-,*,/) & 1000 & \cite{patel-etal-2021-nlp} \\
    \hline
        Dolphin18k & Multi-equation & (+,-,*,/) & 18k & \cite{huang-etal-2016-well} \\
        AQuA-RAT & Multiple-choice & - & 100k & \cite{ling2017program} \\
        Math23k* & Single-equation & (+,-,*,/) & 23k & \cite{huang2017learning} \\
        MathQA & Single-equation & (+,-,*,/) & 35k & \cite{amini-etal-2019-mathqa} \\
        HMWP* & Multi-equation & (+,-,*,/) & 5k & \cite{qin-etal-2020-semantically}\\
        Ape210k* & Single-equation & (+,-,*,/) & 210k & \cite{mwp-bert} \\
        GSM8k & Single-equation & (+,-,*,/) & 8.5k & \cite{gsm8k}\\
        CM17k* & Multi-equation & (+,-,*,/) & 17k & \cite{qin-etal-2021-neural}\\
        \hline
        
    \end{tabularx}
    \caption{Datasets\\(\scriptsize{*Chinese Datasets})}
    \label{tab:datasets}
\end{table}

\subsection*{Small Datasets}
The pioneering work in solving word problems \cite{kushman2014learning}, introduced a classical dataset (Alg514) of 514 word problems, across various domains in algebra (such as percentages, mixtures, speeds etc). This dataset was annotated with multiple equations per problem. AddSub was introduced in \cite{hosseini2014}, with simple addition/subtraction problems, exhibiting limited language complexity. SingleOp \cite{roy2015reasoning} and MultiArith \cite{roy2016solving} were proposed such that there is a control over the operators (single operator in the former and two operators in the latter). SingleEq \cite{koncel2015parsing} is unique in incorporating long sentence structures for elementary level school problems. AllArith \cite{roy_solving} is a subset of the union of AddSub, SingleEq and SingleOp. "Perturb" is a set of slightly perturbed word problems of AllArith, whereas Aggregate is the union of AllArith and Perturb. MAWPS (A \textbf{Ma}th \textbf{W}ord \textbf{P}roblem \textbf{S}olving Repository) \cite{mawps} is a curated dataset (with deliberate template overlap control) that comprises all proposed datasets till that date.  A single equation subset of MAWPS (AsDIV-A) \cite{asdiva} has been studied , for diagnostic analysis of solvers. Similarly, the critique offered by \cite{patel-etal-2021-nlp} was demonstrated using their newly proposed dataset SVAMP. In SVAMP, minutely perturbed word problems from the popular dataset AsDIV-A. This particular subset is used to demonstrate that, while high values of accuracy can be obtained on AsDIV-A easily, SVAMP poses a formidable challenge to most solvers, as it captures nuances in the relationship between similar language formation and dissimilar equations. All aforementioned datasets incorporate an annotation of both the \textit{equation} and the \textit{answer}. Given the subset-superset relationships between some of these datasets, empirical usage of these datasets would need to ensure careful sampling to creating subsets for training, testing and cross-validation. 

\subsection*{Large Datasets}
Dolphin18k \cite{huang-etal-2016-well} is an early proprietary dataset that was evaluated primarily with the statistical solvers. AQuA-RAT \cite{ling2017program} introduced the first large crowd-sourced dataset for word problems with \textit{ rationales} or \textit{explanations}. This makes the setting quite different from the aforementioned datasets, not only with respect to size, but also in the wide variety of domain areas (spanning physics, algebra, geometry, probability etc). Another point of difference is that the annotation involves the entire textual explanation, rather than just the equations. MathQA \cite{amini-etal-2019-mathqa} critically analysed AQuA-RAT and selected the core subset and annotated it with a predicate list, to widen the remit of its usage. Once again, researchers must be mindful of the fact that MathQA is a subset of AQuA-RAT. GSM8k \cite{gsm8k} is a recent single-equation dataset, that is the large scale version of AsDIV-A \cite{asdiva}. Math23K is a popular Chinese dataset for single equation math word problem solving. A recent successor is Ape210k \cite{mwp-bert}.

\section*{Evaluation Measures} 
The most popular metric is \textit{answer accuracy}, which evaluates the predicted equation and checks whether it is the same as the labelled one. The other metric is \textit{equation accuracy}, which predominantly does string matching and assesses the match between the produced equation and the equation from the annotation label.

\section*{Performance of Deep Models}
In this section, we describe the performance of neural solvers towards providing the reader with a high-level view of the comparative performance across the several proposed models.

We have listed the performance of the deep models in Table ~\ref{tab:performance}, on two major datasets - Math23K and MAWPS. 

Some of these deep models report scores on other datasets as well. For conciseness, we have chosen the most popular datasets for deep models. We see that, in general, the models achieve around 70-80 percentage points on \textit{answer accuracy}. \cite{shen-etal-2021-generate-rank} outperforms all other models on Math23k whereas RPKHS \cite{yu-etal-2021-improving} is the best model for MAWPS till date. As mentioned before, graph based models are both popular and effective. A note of caution is that, as inferred from the discussion on datasets, (a) both Math23k and MAWPS are single equation datasets and (b) though some lexical overlap has been performed in the design of these two datasets, the semantic quality of these datasets are quite similar. This aspect has also been experimented and explored in \cite{patel-etal-2021-nlp}. Hence, though we present the best performing algorithms in this table, more research is required to design a suitable metric or a suitable dataset, such that one can conclusively compare these various algorithms.

\begin{table}[h]
    \centering
    \def\arraystretch{1.1}
    \begin{tabular}{l c c c l}
    \hline
        \textbf{Model} & \textbf{Type} & \textbf{AQuA-RAT} & \textbf{MathQA} & \textbf{Source} \\
        \hline
         AQuA & Seq2Seq & 36.4 & - & \cite{ling2017program}\\
         Seq2Prog & Seq2Seq & \textbf{37.9} & 57.2 & \cite{amini-etal-2019-mathqa} \\
         BERT-NPROP & Transformer & 37.0 & - & \cite{piekos-etal-2021-measuring} \\
         Graph-To-Tree & Graph-based & - & \textbf{69.65} & \cite{li-etal-2020-graph-tree} \\
         \hline
    \end{tabular}
    \caption{Performance on Large Multi-Domain Datasets}
    \label{tab:multi-performance}
\end{table}
Apart from these algebraic datasets, multi-domain datasets MathQA and AquA are also of special interest. This is described in Table ~\ref{tab:multi-performance}. The interesting takeaway is that, the addition of BERT modelling to AQuA \cite{piekos-etal-2021-measuring}, still performed slightly worse than the Seq2Prog \cite{amini-etal-2019-mathqa} model, which is a derivative of the Seq2Seq paradigm. 

\begin{table}[h]
    \centering
    \def\arraystretch{1.3}
    \begin{tabularx}{\textwidth}{l l c c l}
    \hline\hline
    Model Name	& Type & Math23k &	MAWPS & Source\\
    \hline
    GTS	& Graph-based & 74.3	& - & \cite{tree-ijcai2019} \\
    SAU-SOLVER & Graph-based & 74.8 &	- & \cite{chiang-chen-2019-semantically} \\
    Group-att & Transformer & 69.5 & 76.1 & \cite{li-etal-2019-modeling}\\
    Graph2Tree & Graph-based & 77.4 & - & \cite{li-etal-2020-graph-tree} \\
    KA-S2T	& Graph-based & 76.3 & - & \cite{wu-etal-2020-knowledge} \\	
    NS-Solver	& Seq2Seq & 75.67 & - & \cite{qin-etal-2020-semantically} \\	
	Graph-To-Tree & Graph-based & 78.8 & - & \cite{li-etal-2020-graph-tree}\\
    TSN-MD	& Teacher Student & 77.4 & 84.4 & \cite{ijcai2020-555}\\
    Graph-Teacher & Graph \& Teacher &	79.1 & 84.2 & \cite{ijcai2021-485}\\
    NumS2T	& Graph-based & 78.1 & - & \cite{wu-etal-2020-knowledge} \\	
    Multi-E/D & Graph-based & 78.4 & - & \cite{shen-jin-2020-solving} \\	
    EPT	& Transformer & - & 84.5 & \cite{kim-etal-2020-point}\\ 
    Seq2DAG	& Graph-based & 77.1 & - & \cite{Cao_Hong_Li_Luo_2021} \\	
    EEH-D2T	& Graph-based & 78.5 & 84.8 & \cite{wu-etal-2021-edge-enhanced} \\
    Generate and Rank	& Graph-based & \textbf{85.4} & 84.0 & \cite{shen-etal-2021-generate-rank} \\
    HMS	& Graph-based & 76.1	& 80.3 & \cite{hms} \\
    RPKHS	& Graph-based & 83.9 & \textbf{89.8} & \cite{yu-etal-2021-improving}\\
    CL & Contrastive Learning & 83.2 & - & \cite{contrastive} \\
    GTS+RODA & Graph-based & 77.9	& - & \cite{augmentation2022}\\
    \hline\\
    \end{tabularx}
    \caption{Answer Accuracy of Deep Models}
    \label{tab:performance}
\end{table}

\section*{Analysis of Deep Models}
In this section of the paper, we analyze the pros and cons of applying deep learning techniques to solve word problems automatically. At the outset, two layers of understanding are imperative: (i) linguistic structures that describe a situation or a sequence of events and (ii) mathematical structures that govern these language descriptions. Though deep learning models have rapidly scaled and demonstrated commendable results for capturing these two characteristics, a closer look reveals much potential for further exploration. The predominant modus-operandus is to create a deep model that converts the input natural language to the underlying equation. In some cases, the input is converted into a set of predicates \cite{amini-etal-2019-mathqa} or explanations \cite{ling2017program}.

\subsection*{What Shortcuts are being Learned?}
Shortcut Learning \cite{shortcut-learning} is a recently well-studied phenomenon of deep neural networks. It describes how deep learning models learn patterns in a shallow way and fall prey to questionable generalizations across datasets (an example is an image being classified as sheep if there was grass alone; due to peculiarities in the dataset).
This is a function of the low-level input we provide to such models (pixels, word embeddings etc.). In the context of word problems, \cite{patel-etal-2021-nlp} exposed how removing the question and simply passing the situational context, leads to the correct equation being predicted. This suggests two things, issues with model design as well as issues with dataset design. The datasets have high equation template overlap, as well as text overlap. Word problem solving is a hard because two otherwise identical word problems, with a small word change (say changing the word \textit{give} to \textit{take}), would completely change the equation. Hence high lexical similarity does not translate to corresponding similarity in the mathematical realm \cite{patel-etal-2021-nlp,sundaram2020distributed}, and attention to key aspects within the text is critical. 

\subsection*{Is Language or Math being Learned?}
\begin{table}[h]
    \centering
    \footnotesize
    \def\arraystretch{1.5}
    \begin{tabular}{|p{5.1cm} | c|}
    \hline
    \textbf{Problem} & \textbf{Solved?} \\
    \hline
    \hline
    John has 5 apples. Mary has 2 apples more than John. How many apples does Mary have? & Yes \\
    \hline
    John has 5 apples. Mary has 2 apples more than John. Who has less apples? & No \\
    \hline
    What should be added to two to make it five? & No \\
    \hline
    \end{tabular}
    \caption{Behaviour of Baseline BERT Model}
    \label{tab:sota}
\end{table}

The question that looms large is whether adequate mapping of language to math has been modelled, whether linguistic modelling has been unfavourably highlighted or that the mathematical aspects have been captured succinctly. We observe that there are opportunities to refine the modelling of both language and math aspects of word problems. Apart from the perturbations experiment done by SVAMP \cite{patel-etal-2021-nlp}, which exposes that the mapping between linguistic and mathematical structures is not captured, we suggest two more experimental analysis frameworks that illustrate deficiencies in linguistic and mathematical modelling. The first one involves imposing a question answering task on top of the word problem as a probing test. For example, a baseline BERT model that converts from input language to equation (Table ~\ref{tab:sota}), trained on MAWPS, can solve a simple word problem such as "\textit{John has 5 apples. Mary has 2 apples more than John. How many apples does Mary have?}", but cannot answer the following allied question "\textit{John has 5 apples. Mary has 2 apples more than John. Who has less apples?}". One reason is of course, dataset design. The governing equation for this problem is "X = 5-2". However, the text version of this, "\textit{What should be added to two to make it five?}", cannot be solved by the baseline model. Similarly, many solvers wrongly output equations such as "X = 2 - 5" \cite{patel-etal-2021-nlp}, which suggests mathematical modelling of subtraction of whole numbers could potentially be improved by simply embedding more basic mathematical aspects. Hence, we observe, that deep translation models neither model language, nor the math sufficiently.

\subsection*{Is Accuracy Enough?}
As suggested by the discussion above, a natural line of investigation is to examine the evaluation measures, and perhaps the error measures for the deep models, in order to bring about a closer coupling between syntax and semantics. High accuracy of the models to predicting the answer or the equation suggests a shallow mapping between the text and the mathematical symbols. This is analogous to the famously observed McNamara fallacy\footnote{\url{https://en.wikipedia.org/wiki/McNamara_fallacy}}, which cautions against the overuse of a single metric to evaluate a complex problem. One direction of exploration is data augmentation with a single word problem annotated with multiple equivalent equations. Metrics that measure the soundness of the equations generated, the robustness of the model to simple perturbations (perhaps achieved using a denoising autoencoder) and the ability of the model to discern important entities in a word problem (perhaps using an attention analysis based metric), are the need of the future. An endeavour has been done by \cite{kumar-etal-2021-adversarial-examples}, where adversarial examples have been generated and utilised to evaluate SOTA models.

\subsection*{Are the Trained Models Accessible?}
Most of the SOTA systems come with their own, well-documented repositories. Though an aggregated toolkit \cite{lan2021mwptoolkit} (open-source MIT License) is available, running saved models in inference mode, to probe the quality of the datasets, proved to be a hard task, with varying missing hyper-parameters or missing saved models. This, however, interestingly suggests that API's that can take a single word problem as input and computes the output, would be highly useful for application designers. This has been done in the earlier systems such as \cite{roy2018mapping} and \cite{wolfram2015wolfram}. 

\section*{Analysis of Benchmark Datasets}
In this section of the paper, we explore the various dimensions of the popular datasets (Table ~\ref{tab:datasets}) with a critical and constructive perspective.

\subsection*{Low Resource Setting}
Compared to usual text related tasks, the available datasets are quite small in size. They also suffer from a large lexical overlap \cite{amini-etal-2019-mathqa}. This taxes algorithms, that now have to generalise from an effectively small dataset. The fact that the field of word problem solving is niche, where we cannot simply lift text from generic sources like Wikipedia, is one of the primary reasons why these datasets are small. Language precision is required, while maintaining mathematical sense. Hence, language generation is also a hard task.

\subsection*{Annotation Cost}
The datasets currently have little to no annotation costs involved as they are usually scraped from homework websites. There are some exceptions that involve crowd-sourcing \cite{ling2017program} or intermediate representations apart from equations \cite{amini-etal-2019-mathqa}. 

\subsection*{Template Overlap}
Many studies \cite{survey2} have demonstrated that there is a high lexical and mathematical overlap between the word problems in popular datasets. While lexical overlap is desirable in a principled fashion, as demonstrated by \cite{patel-etal-2021-nlp}, it often limits the diversity and thus utility of the datasets. Consequently, many strategies have been adopted to mitigate such issues. Early attempts include controlling linguistic and equation template overlap (\cite{mawps}, \cite{asdiva}). Later ideas revolve around controlled design and quality control of crowd-sourcing \cite{amini-etal-2019-mathqa}.

\section*{Road Ahead}
In this section, we describe exciting frontiers of research for word problem solving algorithms.

\subsection*{Semantic Parsing}
As rightly suggested by \cite{survey2}, the closest natural language task for word problem solving is that of \textit{semantic parsing}, and not \textit{translation} as most of the deep learning models have modelled. The mapping between extremely long chunks of text to short equation sentences has the advantage of generalising on the decoder side, but equally has the danger of overloading many involved semantics into a simplistic equation model. To illustrate, an equation may be derived after applying a sequence of steps that is lost in a simple translation process. A lot of efforts have already been employed in adding such nuances in the modelling. One way is to model the input intelligently (for e.g., \cite{mwp-bert}) Here, sophisticated embeddings are learned from BERT based models, using the word problem text as a training bed. The intermediate representations include simple predicates \cite{roy2018mapping}, while others involve a programmatic description (\cite{ling2017program}, \cite{amini-etal-2019-mathqa}). Yet another way is to include semantic information in the form of graphs as shown in (\cite{huang-etal-2018-using}, \cite{chiang-chen-2019-semantically}, \cite{qin-etal-2020-semantically}, \cite{li-etal-2020-graph-tree},  etc.)).

\subsection*{Informed Dataset Design}

As most datasets are sourced from websites, there is bound to be repetition. Efforts invested in modelling things such as the following could help aiding word problem research: (a) different versions of the same problem, (b) different equivalent equation types, (c) semantics of the language and the math. A step in this direction has been explored by \cite{patel-etal-2021-nlp}, which provides a challenge dataset for evaluating word problems, and \cite{kumar-etal-2021-adversarial-examples} where adversarial examples are automatically generated. 

\subsubsection*{Dataset Augmentation}
A natural extension of dataset design, is dataset augmentation. Augmentation is a natural choice when we have datasets that are small and focused on a single domain. Then, linguistic and mathematical augmentation can be automated by domain experts. While template overlap is a concern in dataset design, it can be leveraged in contrastive designs as in \cite{sundaram2020distributed,contrastive}. A principled approach of reversing operators and building equivalent expression trees for augmentation has been explored here \cite{augmentation2022}.

\subsubsection*{Few Shot Learning}
This is useful if we have a large number of non-annotated word problems or if we can come up with complex annotations (that capture semantics) for a small set of word problems. In this way \textit{few shot learning} can generalise from few annotated examples.

\subsection*{Knowledge Aware Models}
We propose that word problem solving is more involved than even semantic parsing. From an intuitive space, we learn language from examples and interactions but we need to be explicitly \textit{trained} in math to solve word problems \cite{marshall}. This suggests we need to include mathematical models into our deep learning models to build generalisability and robustness. As mentioned before, a common approach is to include domain knowledge as a graph~\cite{chiang-chen-2019-semantically,wu-etal-2020-knowledge,qin-etal-2020-semantically,qin-etal-2021-neural}.

\section*{Conclusion}
In this paper, we surveyed the existing math word problem solvers, with a focus on deep learning models. Deep models are predominantly modeled as encoder-decoder models, with input as text and decoder output as equations. We listed several interesting formulations of this paradigm - namely Seq2Seq models, graph-based models, transformer-based models, contrastive models and teacher-student models. In general, graph based models tend to capture complex structural elements that can benefit both linguistic and mathematical aspects. We then explored in detail the various datasets in use. Subsequently, we analysed the various approaches of modelling word problem solving, followed by the characteristics of the popular datasets. We saw an overwhelming trend that paying heed to the mathematical modelling and tying to the linguistic aspects reaped rich dividends. We concluded that the brittleness of the SOTA models was due to: (a) modelling decisions, and (b) dataset design.  This is intended as a comprehensive survey, but the authors acknowledge that there may be methods that have escaped their attention. We also caution that the analysis provided could be subjective and opinionated, and there could be legitimate disagreements with the perspectives put forward. Finally, we mentioned few avenues of further exploration such as the use of semantically rich models, informed dataset design and incorporation of domain knowledge.

\bibliographystyle{apa}
\bibliography{refs}
\end{document}